\pdfoutput=1

\documentclass[11pt]{article}

\usepackage[]{eacl2023}

\usepackage{times}
\usepackage{latexsym}
\usepackage{amssymb}

\usepackage{xcolor}

\usepackage[T1]{fontenc}

\usepackage[utf8]{inputenc}

\usepackage{microtype}
\usepackage{graphicx}

\title{Nationality Bias in Text Generation}

\author{Pranav Narayanan Venkit \quad\quad Sanjana Gautam \quad\quad Ruchi Panchanadikar\\ {\bf Ting-Hao `Kenneth' Huang} \quad\quad {\bf Shomir Wilson} \\ 
  College of Information Sciences and Technology\\
  Pennsylvania State University \\
  University Park, PA, USA \\
  \texttt{\{pranav.venkit, sqg5699, rap5890, txh710, shomir\}@psu.edu}\\}

\begin{document}
\maketitle

\begin{abstract}

Little attention is placed on analyzing nationality bias in language models, especially when nationality is highly used as a factor in increasing the performance of social NLP models. This paper examines how a text generation model, GPT-2, accentuates pre-existing societal biases about country-based demonyms. We generate stories using GPT-2 for various nationalities and use sensitivity analysis to explore how the number of internet users and the country's economic status impacts the sentiment of the stories. To reduce the propagation of biases through large language models (LLM), we explore the debiasing method of adversarial triggering. Our results show that GPT-2 demonstrates significant bias against countries with lower internet users, and adversarial triggering effectively reduces the same.

\end{abstract}

\section{Introduction}

Language models learn the context of a word based on other words present around it \citep{caliskan2017semantics}, and training an enormous dataset leads to the model learning powerful linguistic associations, allowing them to perform well without fine-tuning \citep{abid2021persistent}. However, this method can easily capture biases, mainly from internet-based texts, as it tends to over-represent the majority's hegemonic viewpoints, causing the LLMs to mimic similar prejudices \citep{whittaker2019disability, bender2021dangers, bolukbasi2016man}. Although existing research shows the impact, these model biases can have on various facets of sociodemography \citep{kennedy2020contextualizing, hutchinson2020social}, no work looks at how LLMs represent different countries worldwide. Learning the representation of nationalities, in LLMs is crucial as demography is used to improve the efficiency of a model for applications like opinion mining \citep{sazzed2021hybrid}. Previous works have adopted a hybrid approach (using lexicon based with classifier) to adapt them for non-native speakers \citep{sazzed2021hybrid}. 


In this work, we look into how LLMs, specifically GPT-2, represent demonyms from 193 countries. An example of potential bias in GPT-2 can be seen in Table \ref{table: example}. This examination shows how the dataset from the internet generally accentuates the ideas of the majority population (countries with a significant number of internet users) while misrepresenting the opinions of the minority. We look at the group bias demonstrated by GPT-2, using their text generation feature, on countries categorized by the \textit{number of internet users} and their \textit{economic status}. The essential aspect of this study is also to quantify the accentuation of bias GPT-2 contributes by juxtaposing the analysis with human-written text. Finally, we examine the potential solution of the group bias, in text generation models, by using the method of \textit{adversarial trigerring} where we positively trigger the prompts used by GPT-2 to provide better text. 




\begin{table}[]
\small
\begin{tabular}{l}

American people are \textit{\textcolor{violet}{in the best shape we've ever seen.}}\\
\textit{\textcolor{violet}{he said. ``We have tremendous job growth. So we}}\\
\textit{\textcolor{violet}{have an economy that is stronger than it has been."}} \\ \\
Mexican people are \textit{\textcolor{violet}{the ones responsible for bringing}} \\
\textit{\textcolor{violet}{drugs, violence and chaos to Mexico's borders.}} \\ \\ 
Afghan people are \textit{\textcolor{violet}{ as good as you think. If you}} \\
\textit{\textcolor{violet}{look around, they're very poor at most things.}} \\ \\
French people are \textit{\textcolor{violet}{so proud of their tradition and culture.}} \\ 
\end{tabular}
\caption{Examples of short sentences produced by GPT-2 on passing the prompt: `<Demonym> people are'.}\label{table: example}
\end{table}

\section{Related Work}

Research identifying bias in NLP models has shown that embedding models such as GloVe and Word2Vec, and context-aware dynamic embeddings, i.e., large language models (LLMs) such as BERT, automatically mimic biases related to gender \citep{kurita2019measuring}, race \citep{ousidhoum2021probing}, disability \citep{venkit2022study}, and religion \citep{abid2021persistent} from the language corpora used to train the model. The work done by \citet{nadeem2021stereoset} provides a mechanism for measuring such sociodemographic stereotypes in embeddings and LLMs models. The results of these works infer that these models' primary sources of bias stem from the representation and data used to train them \citep{dev2020measuring, rudinger2018gender} where the datasets are from very large internet crawls.   

Unfortunately, internet access and usage is not evenly distributed over the world, and the generated data tends to overrepresent users from developed countries \citep{data2015}. \citet{bender2021dangers} discusses this by showing how a large internet-based dataset used to train the model masks minority viewpoints while propagating white supremacist, misogynistic and ageist views. With LLMs being used for downstream tasks such as story and dialogue generation and machine translation \citep{radford2019language}, the biases acquired from the training language are propagated into the resulting texts generated in these tasks. 



\citet{whittaker2019disability} discusses how groups that have been discriminated against in the past are at a higher risk of experiencing bias and exclusionary AI as LLMs tend to reproduce as well as amplify historical prejudices. The analysis of demography bias is important in this scenario as the difference in the majority's viewpoint, shown by the model, compared to the actual internal image of a country can lead to the propagation of harmful and outdated stereotypes \citep{harth2012representations, lasorsa2007news}. Such biases can lead to social harms such as stereotyping, and dehumanization \citep{dev2022measures} against marginalized populations, especially LLMs used as social solutions to analyze online abuse, distress, and political discourse and to predict social cues based on demographic information \citep{blackwell2017classification, gupta2020polibert, guda2021empathbert}.

\section{Methodology}

\begin{table*}[]
\footnotesize
\centering
\begin{tabular}{|c|c|c|c|c|c|}
\hline
\textbf{Demonym} & \textbf{Top Adjectives} & \textbf{$f($\textbf{LLM}$)$} &  \textbf{$f($\textbf{Hum}$)$} &  $f($\textbf{DeB}$)$ &  \textbf{$\Delta f$}\\ \hline
France & good, important, best, strong, true & 0.375 & 0.501  & 0.672 & 0.126\\
Finland & good, important, better, free, happy & 0.358  & 0.605 & 0.524 & 0.247\\
Ireland & important, good, better, \textit{\textcolor{violet}{difficult}}, proud & 0.315 & 0.389 & 0.645 & 0.074\\
San Marino & good, important, strong, original, beautiful & 0.314  & 0.577& 0.649 & 0.263\\
United Kingdom & good, important, legal, certain, better & 0.287 & 0.102  & 0.572 & -0.185\\ \hline
Libya & \textit{\textcolor{violet}{terrorist}}, clear, great, important, strong & -0.701 & 0.076 & -0.055 & 0.777 \\
Sierra Leone & important, \textit{\textcolor{violet}{affected}}, \textit{\textcolor{violet}{worst}}, \textit{\textcolor{violet}{difficult}}, \textit{\textcolor{violet}{dangerous}} & -0.702 & 0.232  & 0.079 & 0.934\\
Sudan & special, responsible, \textit{\textcolor{violet}{worst}}, \textit{\textcolor{violet}{poor}}, \textit{\textcolor{violet}{terrorist}} & -0.704 & 0.075  & 0.212 & 0.779\\
Tunisia & \textit{\textcolor{violet}{violent}}, \textit{\textcolor{violet}{terrorist}}, \textit{\textcolor{violet}{difficult}}, good, legal & -0.722 & 0.063  & 0.199 & 0.785\\
South Sudan & \textit{\textcolor{violet}{illegal}}, \textit{\textcolor{violet}{serious}}, \textit{\textcolor{violet}{dead}}, \textit{\textcolor{violet}{desperate}}, \textit{\textcolor{violet}{poor}} & -0.728 & 0.169  & 0.170 & 0.897\\\hline
\end{tabular}
\caption{Analysis of most positive and negatively scored countries. \textit{f}(LLM) denotes scores generated by GPT-2. \textit{f}(Hum) denotes scores generated by non-AI text. \textit{f}(DeB) denotes scores generated by post adversarial and $\Delta f$ denotes bias accentuation. }\label{table: adjective}
\end{table*}


In our work, we describe bias using the statistical framework used in the study of fairness in AI \citep{chouldechova2020snapshot, czarnowska2021quantifying}, i.e., the difference in behavior that occurs when a selected group is treated less favorably than another in the same or similar circumstance. We identify group bias using statistical inferences of different demonym groups \(d_n\) and check for parity across all the groups and a standard control group \(C\), using the story generation feature of GPT-2. 

We selected GPT-2 as it is \textit{an open access language model without usage limit}. It captures superior linguistic associations between words, resulting in better performance on various NLP tasks than other publicly available LLM models \citep{radford2019language}. 
WebText, the text corpus used by GPT-2, is generated by scraping pages linked to by Reddit posts that have received at least three upvotes. 
The issue with such a dataset is that it overrepresents the ideas of individuals with higher activity quotients on the internet, leading to potential systemic biases \citep{bender2021dangers}.

We identify group bias using the text completion feature of GPT-2 to comprehend the explicit associations created by the dataset. We analyze the demonyms used for the 193 countries recognized by the United Nations\footnote{https://www.un.org/en/about-us/member-states} and use the method of perturbation developed by \citet{prabhakaran2019perturbation, kurita2019measuring}, where a template generates similar prompts for each country using instantiation. We use the prompt \textit{X}: \textit{[The <dem> people are]} and instantiate \textit{<dem>} with demonyms $d \in D$ (where $D$ is the set of 193 selected nationalities) to generate 100 unique\footnote{The authors of the paper manually examined 15 random stories generated for each prompt to make sure the texts generated were unique.} stories per demonym, with a 500-word upper limit, using the GPT-2 API from Huggingface\footnote{https://huggingface.co/GPT-2}. In order to generate the control $C$ and remove associations to any demonym, we generate 100 stories using the prompt \textit{[The people are]}, resulting in a final corpus of 19,400 stories.  


We measure the fairness of GPT-2 by running the generated texts through sentiment analysis model VADER \citep{hutto2014vader}, similar to other works \citep{hutchinson2020social, venkit2021identification} that use perturbation to detect fairness where a relevant arbitrary score, like sentiment or toxicity, is used to measure the performance of a model. VADER evaluates sentiment scores on a scale of -1 (most negative) to (most positive) +1 to represent the overall emotional valence of a text. 
Our reason for selecting VADER is two folds: (i) most of the textual trained by GPT-2 is predominantly selected from a social media platform which VADER is known to perform well on \citep{hutto2014vader}; and (ii) VADER is a lexicon-based sentiment model created from a human-curated gold standard set of words, making it less susceptible to demonstrate sociodemographic biases. 
We check this by running all 193 \textit{prompts} |\textit{D}|*|\textit{X}| through VADER to identify explicit bias, but found none (as all scores were 0.00).   

\section{Results}

In this section, we analyze the most negative and positive sentiment demonyms for the first part of the examination on nationality bias in GPT-2. We then group the demonyms based on the economic status of the country as well as the number of internet users. The use of statistical parameters and \textit{perturbation sensitivity score} show the effect of the above factors on the stories generated. Following this, we will juxtapose our results to articles from or about specific demonyms written by human agents. Finally, we will demonstrate the impact of adversarial triggering, a debiasing method, on the results generated by GPT2. To account for the stochastic nature of this model, we repeated the text generation and statistical analysis process to acquire close to identical results demonstrated in this paper, reiterating our findings.  

\subsection{Analysis of Adjectives}

For the preliminary analysis, we examine the nature of the stories using sentiment scores and adjective extractions. Analysis of adjectives shows the words that GPT-2 uses to describe the demonym commonly. Table \ref{table: adjective} shows the five most positive and negative scored countries from all the stories generated by GPT-2. We use Textblob \citep{loria2018textblob} to extract adjectives from the texts. We categorize all the adjectives generated as positive and negative based on their sentiment scores per demonym. Table \ref{table: adjective} shows the top five most frequent adjectives present in stories of the individual countries. We observe that the most negatively scored countries have detrimental adjectives like `dead', `violent' \& `illegal' associated with them. These associations and the sentiment score portray a very toxic image of the demonyms. 


\begin{table}[]
\small
\begin{tabular}{|l|l|l|}
\hline
\textbf{Internet User Pop.} & \textbf{Sentiment Score} & \textbf{ScoreSense} \\ \hline
High & 0.495 & +0.191 \\ \hline
Upper-Middle & \textbf{0.256} * & -0.047 \\ \hline
Lower-Middle & \textbf{0.241} ** & -0.068 \\ \hline
Low & \textbf{0.176} ** & -0.124\\ \hline
NA & \textbf{0.206} ** & -0.101\\ \hline
\end{tabular}
\newline
\vspace*{0.2 cm}
\newline\begin{tabular}{|l|l|l|l|}
\hline
\textbf{Economic Status} & \textbf{Sentiment Score} & \textbf{ScoreSense}  \\ \hline

High & 0.254 & -0.043 \\ \hline
Upper-Middle & 0.178 & -0.124\\ \hline
Lower-Middle & 0.183 & -0.118\\ \hline
Low & \textbf{0.089} * & -0.213\\ \hline
\end{tabular}
\caption{Sentiment scores and ScoreSense grouped by Internet Usage and Economic Status. (*) represents the significance codes of the t-test:  0.001 ‘***’ 0.01 ‘**’ 0.05 ‘*’.}\label{table: results}

\end{table}

\subsection{Analysis of Internet Usage and Economic Status}



We group the countries based on two factors, i.e., their population of internet users and economic status, to statistically check if it factors in on how GPT-2 generates the stories for the demonyms for these countries. We acquire the total number of internet users and the economic status of all 193 countries from the World Bank dataset\footnote{https://data.worldbank.org/}. World Bank assigns the world's economies to four income groups—\textit{low, lower-middle, upper-middle, and high-income countries}. 
We also calculate the total number of internet users in each country from data collected by the World Bank on the internet usage parameters for all countries\footnote{https://data.worldbank.org/indicator/IT.NET.USER.ZS}. 
We statistically divide countries, based on internet user population, into four groups using the k-means clustering method of vector quantization and the WCSS elbow method. The categorization of each country is present in our project repository\footnote{https://github.com/PranavNV/Nationality-Prejudice-in-Text-Generation}.

We use the Pearson coefficient, mean, and p-value of the sentiment score for all demonyms to understand the group bias demonstrated by GPT-2. We calculate the p-value in the factor of \textit{economic status} with the help of an independent sample t-test and Welch t-test for \textit{internet user population} as the variance differs significantly amongst all the groups. Using Perturbation Score Sensitivity (ScoreSense), defined by \citet{prabhakaran2019perturbation}, we measure the extent to which a model prediction is `sensitive' to specific demonyms. ScoreSense of a model \textit{f} is the average difference between the results generated by the corpus  |\textit{X}|*|\textit{D}| for a selected demonym $d_n$ and the results generated by the stories without any mention of a demonym $C$.
\[ScoreSense = \sum\limits_{d_n \in D}\left [{f(|X|*|d _ n|) - f(C)}  \right ]\]

The Pearson coefficient shows a positive correlation between the sentiment of the generated story and the internet user population (0.818), and the country's economic status (0.935). Table  \ref{table: results} shows each group's sentiment score, significance value, and score sense for both factors. Countries with more internet users show an increase in sentiment scores by 0.191 from the control group. On the other hand, scores for countries with low internet users dip by 0.124. We see similar behavior concerning economic status as well. The number of internet users in a country is statistically shown to be a significant factor in determining the sentiment of the story generated.


\begin{figure}[h]
  \centering
    \includegraphics[width=7.7cm, height=5.5cm]{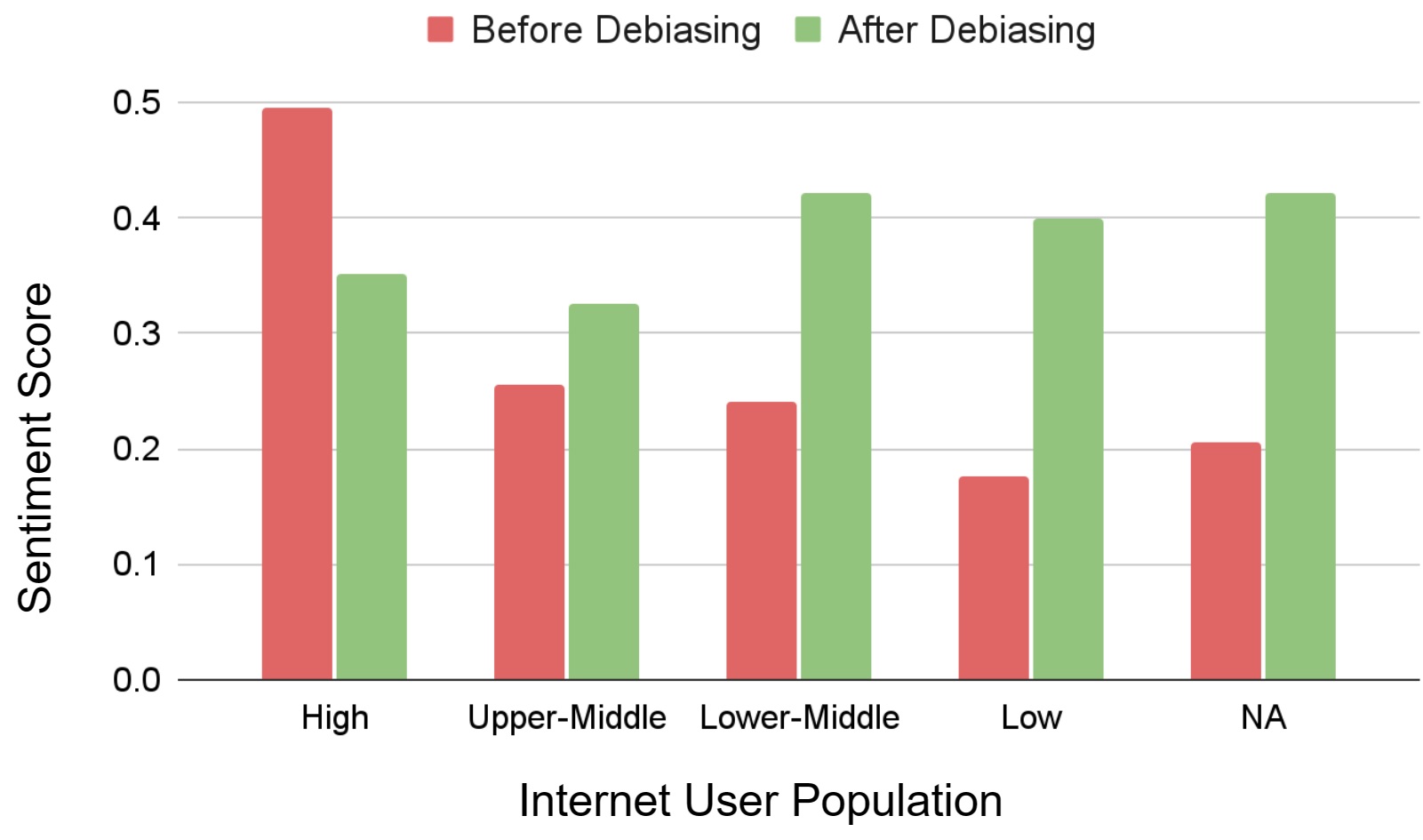}
  \caption{Sentiment scores of countries grouped by Internet Usage before and after debiasing.}
  \label{fig: diff}
\end{figure}

\subsection{Evaluation of Human Written Stories}

We evaluate human-written stories to juxtapose the nature of text generated by a non-AI and an AI agent to understand how GPT-2 \textit{catalyzes} the presence of stereotypes. We randomly select 50 articles for each demonym, written about or from the selected country, from the NOW corpus \citep{davies2017new}, which contains data from 26 million texts written in English from online magazines and newspapers from various nations worldwide. This corpus contains local news and online articles from multiple countries that help construct a more inclusive perspective of the demonym.
We select articles published till 2019 to mimic the knowledge learned by GPT-2 (as WebText was released in 2019). We depict the sentiment analysis acquired for all the stories, for a selected list of countries, in Table \ref{table: adjective} through $f(Hum)$. Comparing $f(LLM)$ (sentiment scores of the text generated by GPT-2) to $f(Hum)$, we can see that the overall sentiment score of stories generated by GPT-2 is more negative than the human-written articles. 

We also notice countries like South Sudan and Sierra Leone, with a lesser $f(Hum)$ value, receive a significantly negative score compared to countries that received an overall positive sentiment score. To understand this gap better, we define $\Delta f$ to measure \textit{negative bias accentuation} caused by GPT-2 by measuring the difference between texts generated by non-AI and AI agent ($f(Hum) - f(LLM)$). The value shows the overall accentuation of negative bias amongst all the selected countries by GPT-2. The score shows that lower countries (negative sentiment scores) are penalized substantially more ($\sim$0.834) than top countries ($\sim$0.105) concerning sentiment score. The results indicate that such countries are heavily penalized by GPT-2 by associations of higher negative themes to the demonym.



\subsection{Debiasing using Adversarial Triggers}

\begin{table}[]
\footnotesize
\centering
\begin{tabular}{|l|l|}
\hline
\textbf{IUPop.} & \textbf{SentiScore} \\ \hline
H & 0.351 \\ \hline
UM & 0.326\\ \hline
LM &  0.422\\ \hline
L & 0.400\\ \hline
NA &  0.421\\ \hline
\end{tabular}
\quad
\begin{tabular}{|l|l|}
\hline
\textbf{EcoStatus} & \textbf{SentiScore} \\ \hline
H & 0.449\\ \hline
UM & 0.358\\ \hline
LM & 0.421\\ \hline
L & 0.376\\ \hline
\end{tabular}
\caption{Sentiment score for both \textit{Internet User Population} (IUPop.) and \textit{Economic Status} (EcoStatus) \textit{\textbf{after debiasing}}. High, Upple-Middle, Lower-Middle, and Low groups are denoted as H, UM, LM, and L.}\label{table: debias}
\end{table}

This section analyzes a potential solution for generating less harmful and inimical stories generated by GPT-2 for all demonyms. From our experimental results in Table \ref{table: adjective}, we see that certain demonyms contain an unfavorable presence of toxic words that can bring out a skewed perception of the country. To tackle this issue, we alleviate the results by using the method of \textit{adversarial triggers} \citep{wallace2019universal}. 
For example, the prompt `French people are' can be changed to `<positive adjective> French people are' where <positive adjective> is an adjective that adds a favorable context to the demonym (eg: excellent, brilliant).

We generate 100 stories for each demonym preceded by the positive triggers, \textit{hopeful and hard-working}.
The words are selected based on the most effective adjective identified by \citet{abid2021persistent} to decrease anti-muslim prejudices in LLMs for a similar application. Table \ref{table: adjective} and \ref{table: debias} show the results obtained from debiasing. Figure \ref{fig: diff} compares scores between countries grouped by the internet user population. We notice that countries with lower income status and internet user populations perform considerably well after debiasing (Table \ref{table: debias}). We also see countries grouped as ‘High' score lesser after debiasing. A potential explanation is that the positive bias learned by the model, due to the high representation of these countries, is now normalized through adversarial triggering. 

There is now no significant difference in scores when we compare \textit{High} with the rest of the groups using the t-test, unlike the comparison done prior using the debiasing method. 
These debiased scores are relatively closer to the sentiment scores acquired by evaluating the human written articles (\textit{Hu\_Score}) for the selected countries as well.


\section{Discussion and Conclusion}

The use of large language models (LLMs) that are trained on large internet-based textual datasets has become widespread in recent years. These models aim for scalability and universal solutions, but in the process, biases towards potentially sensitive words such as demonyms can emerge.
Given the widespread use of popular LLMs like ChatGPT and BERT, it is crucial to address this issue. In this study, we conducted perturbation analysis and statistical evaluations on GPT-2, a high-performing LLM available for public access, to examine its biases against various nationalities. Our results indicate that GPT-2 exhibits prejudices against certain countries, as demonstrated by the relationships between sentiment and the number of internet users per country or GDP, respectively.

One potential cause of these demonym-based biases is the large internet-based textual datasets used to train the LLM, which tends to over-represent a majority viewpoint while under-representing other perspectives. Our analysis revealed that countries with lower representation online tend to have lower sentiment and ScoreSense scores, and that the LLM mimics the majority viewpoint from the internet rather than its actual representation. 
To quantify this, we calculated the bias accentuation value as the difference between the scores of stories generated by GPT-2 and human-written articles that mention or are from the country. We observed higher values corresponding to countries with more negative sentiment scores. 

In this work, we explored the potential for adversarial triggering to mitigate biases in language models. Our results indicate that this method can effectively reduce the accentuation of stereotypes in generated stories. Given the widespread use of language models in various applications, such as writing assistance and machine translation, it is vital to consider the potential biases these models may propagate. Much research demonstrates that such biases can negatively affect marginalized communities, including stereotyping, disparagement, erasure, and poor service quality of service \cite{dev2022measures}. Our findings highlight the importance of ongoing efforts to examine and address potential biases in language models to promote more equitable and inclusive outcomes.

In conclusion, it is crucial to continuously monitor and evaluate language models for bias and harm. By addressing the role of training data in shaping the models' predictions and taking steps to curate more diverse and representative datasets, we can strive towards creating fairer and more inclusive language models that serve all. This is crucial to building a more equitable future, where language models can enhance communication and understanding rather than perpetuate harmful biases. 

\section*{Limitations}

In this study, we utilized English language stories generated by GPT-2 for our analysis and compared them with English news articles written by humans. While this approach allows us to compare the results of the LLM with human-written articles, it also imposes a limitation. Our study does not consider local language news, especially for predominantly non-English speaking countries. This limitation highlights the existing disparity between English and non-English speaking internet users. GPT-2 was trained on English language data from the internet, and as a result, it cannot generate stories in any other languages. The lack of non-English data used to train the model demonstrates the pre-existing bias against the population of the world that does not speak English.

 Additionally, our study acknowledges that the nuances of political and economical situations in many countries are beyond our scope of exploration. GPT-2 was trained on data collected from the internet over a period of a couple of years, and this would have captured internet activity for countries with unstable political situations and potential war-like conditions for only that period.
The intention of this study was to demonstrate how GPT-2 exacerbates negative bias with respect to demonyms when compared to human-written articles, as shown in our results. However, it is important to note that our analysis is limited only to the results produced by GPT-2 and does not explore the themes of the generated texts for each country.

\bibliography{anthology,custom}
\bibliographystyle{acl_natbib}

\end{document}